\pdfoutput=1

\documentclass[11pt]{article}

\usepackage[preprint]{acl}

\usepackage{times}
\usepackage{latexsym}

\usepackage[T1]{fontenc}

\usepackage[utf8]{inputenc}

\usepackage{microtype}
\usepackage{colortbl}

\usepackage{booktabs}       
\usepackage{amsfonts}       
\usepackage{nicefrac}       
\usepackage{microtype}      
\usepackage{graphicx}
\usepackage{multirow}
\usepackage{enumitem}
\usepackage{wrapfig}
\usepackage{amsmath}

\usepackage{inconsolata}

\usepackage{graphicx}

\definecolor{softgray}{RGB}{242, 242, 242}  
\definecolor{softblue}{RGB}{208, 231, 255}  

\definecolor{skyblue}{RGB}{135, 206, 235}
\definecolor{lightblue}{RGB}{173, 216, 230}
\definecolor{lightyellow}{RGB}{255, 255, 224}
\title{LanP: Rethinking the Impact of Language Priors in Large Vision-Language Models}

\author{Zongyu Wu$^{1}$\thanks{Equal contribution}, Yuwei Niu$^{2*}$, Hongcheng Gao$^{2}$,  Minhua Lin$^{1}$, Zhiwei Zhang$^{1}$, \\
\textbf{Zhifang Zhang$^{2}$, Qi Shi$^{3}$, Yilong Wang$^{1}$, Sike Fu$^{1}$\thanks{Work done during an internship at Penn State University},  Junjie Xu$^{1}$, Junjie Ao$^{4}$,} \\
\textbf{Enyan Dai$^{1}$, Lei Feng$^{2}$, Xiang Zhang$^{1}$, Suhang Wang$^{1}$}\thanks{Corresponding Author} \\
        \textsuperscript{1}The Pennsylvania State University\\
        \textsuperscript{2}Singapore University of Technology and Design
 \\
 \textsuperscript{3}Peking University
  \textsuperscript{4}Rensselaer Polytechnic Institute\\
         \texttt{zongyuwu@psu.edu, niuyuwei04@gmail.com, szw494@psu.edu} \\
}

\begin{document}
\maketitle
\begin{abstract}
Large Vision-Language Models (LVLMs) have shown impressive performance in various tasks. However, LVLMs suffer from \emph{hallucination}, which hinders their adoption in the real world. Existing studies emphasized that the strong language priors of LVLMs can overpower visual information, causing hallucinations. However, the \emph{positive role of language priors} is the key to a powerful LVLM. If the language priors are too weak, LVLMs will struggle to leverage rich parameter knowledge and instruction understanding abilities to complete tasks in challenging visual scenarios where visual information alone is insufficient. Therefore, we propose a benchmark called \textbf{LanP} to rethink the impact of \textbf{Lan}guage \textbf{P}riors in LVLMs. It is designed to investigate how strong language priors are in current LVLMs. LanP consists of 170 images and 340 corresponding well-designed questions. Extensive experiments on 25 popular LVLMs reveal that many LVLMs' language priors are not strong enough to effectively aid question answering when objects are partially hidden. Many models, including GPT-4 Turbo, exhibit an accuracy below 0.5 in such a scenario.
\end{abstract}

\section{Introduction}
\label{sec:intro}
Large Language Models (LLMs)~\citep{touvron2023llama,touvron2023llama2,jiang2024mixtral,jiang2023mistral,vicuna2023} have been successfully applied in many fields~\cite{lin2024decoding,schmirler2024fine,MaWGSTRGM24,brahmavar2024generating}. The success of LLMs has also advanced the development of Large Vision-Language Models (LVLMs)~\citep{yin2023survey,LiuLWL23a,reid2024gemini,liu2023improved,NEURIPS2023_9a6a435e,fu2023challenger,tong2024cambrian,yao2024minicpm}.
LVLMs have shown impressive performance in producing texts given textual and visual inputs, facilitating various tasks such as visual question answering~\citep{shao2023prompting,sammani2022nlx,fu2023mme} and graph learning~\cite{xu2024llm}. Despite advancements of LVLMs, similar to LLMs~\citep{huang2023survey,ji2023survey}, they also suffer from 
\textit{hallucination}~\citep{DaiLJSF23}, a phenomenon that LVLMs occasionally generate outputs that appear reasonable but are inconsistent with the actual visual content~\citep{bai2024hallucination}. 
This degrades people's trust in LVLMs and hinders their adoption in high-stakes scenarios such as healthcare~\citep{LiWZULYNPG23,pmlr-v225-moor23a} and finance~\citep{bhatia-etal-2024-fintral,wimmer2023leveraging}.

\begin{figure*}[t!] 
\centering 
\includegraphics[width=1\textwidth]{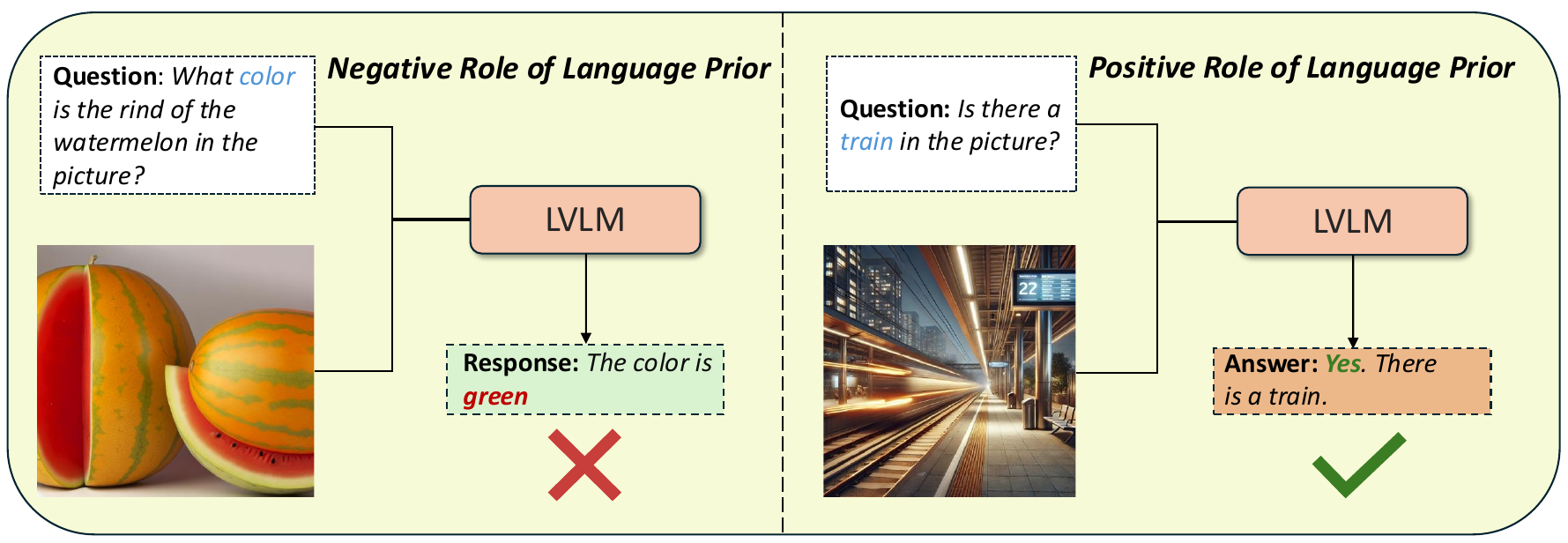}

\caption{An illustration of different roles of language priors in LVLMs. The left half of the image shows an example where language priors have a negative impact, while the right half shows an example where language priors bring a positive impact.}
\label{fig:intro}
\end{figure*}

To evaluate the hallucination in LVLMs, several benchmarks have been developed~\citep{LiDZWZW23,liu2024phd,liu2023hallusionbench,jiang2024hal,zhang2024benchmarking}. Some works~\citep{liu2024phd,liu2023hallusionbench,leng2023mitigating} claim that language priors in LLMs, i.e., the parametric knowledge learned from abundant training data, could negatively affect LVLMs as the strong language bias might overshadow the visual information, causing hallucinations. For example, as shown in the left part of Figure~\ref{fig:intro},  LVLMs might answer that the color of this watermelon's rind is green even if we show them an image where most of the watermelon rind is orange, which is caused by the language priors in LLMs that watermelon's rind is green. One benchmark~\cite{lee2024vlind} also explores scenarios where LVLMs excessively rely on language priors.

However, language priors encompass world knowledge that LLMs learned from extensive training data, which endows LVLMs with the ability to understand information and solve problems. For example, the right part of Figure~\ref{fig:intro} depicts a scenario where a train is in motion, causing its body to appear blurry in the photo. If the model is unable to integrate the entire image scene with its internal parametric knowledge—such as platform information—it may struggle to recognize the train. Hence, language priors are very helpful. They are the key to the success of LVLMs. It is valuable to evaluate how powerful the language priors are in current LVLMs. Therefore, we construct a new benchmark named LanP to rethink the impact of \textbf{Lan}guage \textbf{P}riors in LVLMs. To demonstrate the positive effects of language priors, each image-question pair in LanP is carefully designed. If the model's language priors are weak, it might be hard to rely solely on the visual part to answer the questions;

In summary, our \textbf{main contributions} are: 
\begin{itemize}[leftmargin=1em]
  \item  We meticulously design a benchmark LanP to evaluate the positive role of language priors in LVLMs. LanP consists of 170 images and 340 handcrafted VQA pairs.
  \item Extensive experiments on 25 widely used LVLMs show that our benchmark poses challenges to current LVLMs, indicating that language priors are not powerful enough in LVLMs. For language priors, we should not simply weaken them to avoid hallucinations; instead, we should find an appropriate trade-off where language priors can provide beneficial knowledge while avoiding overriding visual information. 
\end{itemize}

\section{Related work}
\noindent\textbf{Large Vision-Language Models}. 
Benefiting from the combination of LLMs~\citep{radford2018improving,radford2019language,BrownMRSKDNSSAA20,peng2023instruction} with visual modules, LVLMs~\citep{LiuLWL23a,openai2023gpt,claude3,reid2024gemini,liu2023improved,NEURIPS2023_9a6a435e,zhu2023minigpt,chen2024sharegpt4v,ye2023mplug,awadalla2023openflamingo,wang2023cogvlm,bai2023qwen,li2023blip,ye2023mplug,ye2023mplug2,yang2024law,sun2023generative} have acquired strong capability to understand and utilize both visual and textual information, benefiting various tasks. Many open-source LVLMs usually contain three components~\citep{LiDZWZW23}: a vision encoder such as CLIP~\cite{radford2021learning} to encode the vision information, an LLM to provide textual analysis and reasoning, and a cross-modal alignment module to align the information from text modality and vision modality.

\noindent\textbf{Hallucination in LVLMs}. Despite the great performance of LVLMs, they suffer from hallucinations, which generally refer to the phenomenon that the generated text responses are inconsistent with the corresponding visual content~\citep{bai2024hallucination}. Many efforts~\citep{yu2024rlhf,han2024skip,sun2023aligning,liu2023improved,liu2023mitigating,zhai2023halle,chen2023mitigating,yue2024less,huang2023opera,wang2023evaluation,tong2024eyes,jiang2023hallucination,leng2023mitigating,yin2023woodpecker,zhang2024reflective,chen2024ict} have been taken to understand the causes of hallucinations, which can be generally categorized into three categories, i.e., \textit{data-level}, \textit{decoding-level}, and \textit{module-level}. From the \textit{data-level} perspective, the text-image data used for training LVLMs is more challenging to scale than the pure text data for training Large Language Models~\citep{sun2023aligning}. Additionally, data collected from the internet or generated by models often contains noise~\citep{liu2023improved} and lacks diversity~\citep{liu2023mitigating}. Recent studies~\citep{zhai2023halle,chen2023mitigating,yue2024less} have also suggested that overly detailed data, which exceeds the visual perception capabilities of LVLMs, can increase hallucinations. From the \textit{decoding-level} perspective, 
analysis on Opera~\citep{huang2023opera} shows that a model's self-attention might focus more on the previously generated text tokens, thus ignoring the image and causing hallucination. Furthermore, mainstream methods such as top-$K$ sampling, while enhancing the diversity of generated text, also amplify the risk of hallucinations~\citep{wang2023evaluation}. Regarding the \textit{module-level}, some researchers~\citep{tong2024eyes} believe that the perceptual capability of the model's visual encoder is deficient, leading to a loss of visual information during encoding. Weaknesses in the alignment~\citep{jiang2023hallucination} module may also hinder perfect modality alignment. Additionally,~\citet{leng2023mitigating} found that the model's language priors can override the visual content, resulting in hallucinations. Therefore, many studies~\citep{leng2023mitigating,chen2024halc,zhang2024debiasing,deng2024seeing,wu2024noiseboost,xiao2024seeing,sarkar2024mitigating,woo2024don,chen2024alleviating} are dedicated to reducing the influence of language priors.

While this approach enhances the model's performance on benchmarks \citep{liu2023hallusionbench,tong2024eyes,LiDZWZW23} that emphasize visual capability and overlook language priors, we believe that strong language priors are crucial for robust and comprehensive LVLMs. For LVLMs aimed at achieving Artificial General Intelligence, the language priors are not strong enough.

\noindent\textbf{Benchmarks for Evaluating Hallucination}. With the development of research on LVLMs hallucinations, numerous benchmarks~\citep{LiDZWZW23,liu2024phd,liu2023hallusionbench,jiang2024hal,lovenia2023negative,huang2024visual,kaul2024throne,fieback2024metatoken,jing2023faithscore,cui2023holistic,wang2023llm,wang2024mitigating,chen2024unified,cha2024visually,sun2023aligning,zheng2024reefknot,xu2023lvlm,sun2024crosscheckgpt,qu2024look,wang2024understanding} for evaluating hallucinations have been proposed. POPE~\citep{LiDZWZW23} has pioneered the exploration of object hallucinations, discovering that LVLMs tend to generate objects frequently appearing or co-occurring in the training set. They constructed questions based on the frequency of co-occurring words in the corpus, asking LVLMs about the presence of objects. However, with the development of multimodal large models, there is a need for updated and more challenging benchmarks. Hallusionbench~\citep{liu2023hallusionbench} is the first benchmark specifically examining language priors, evaluating the model's ability to handle conflicts between prior knowledge and visual information through manually crafted complex questions.

\noindent\textbf{General Domain LVLMs Benchmarks}. We also introduce some classical general domain LVLMs benchmarks here. General domain LVLM benchmarks aim to holistically measure foundational abilities across vision and language, with examples from early benchmarks like VQAv2~\cite{goyal2017making} and VizWiz~\cite{gurari2018vizwiz}. Representative examples in this category are MME~\cite{abs-2306-13394}, MMBench~\cite{liu2024mmbench}, and SEED-Bench~\cite{li2024seed}, each offering distinct approaches to comprehensive assessment. MME stands out as a benchmark designed to provide a broad evaluation across more than ten perception and cognition tasks, emphasizing real-world scenarios and practical application. MMBench is another comprehensive benchmark, featured by its detailed analysis, allowing for granular insights into model strengths and weaknesses. SEED-Bench and SEED-Bench-2~\cite{abs-2311-17092}, are notable for their large-scale multiple-choice questions spanning a wide range of capabilities. These benchmarks are valuable for measuring overall model performance and tracking progress across various dimensions of multimodal understanding.

\section{The constructed benchmark - LanP}

\begin{figure*}[t!] 
\centering 
\includegraphics[width=1\textwidth]{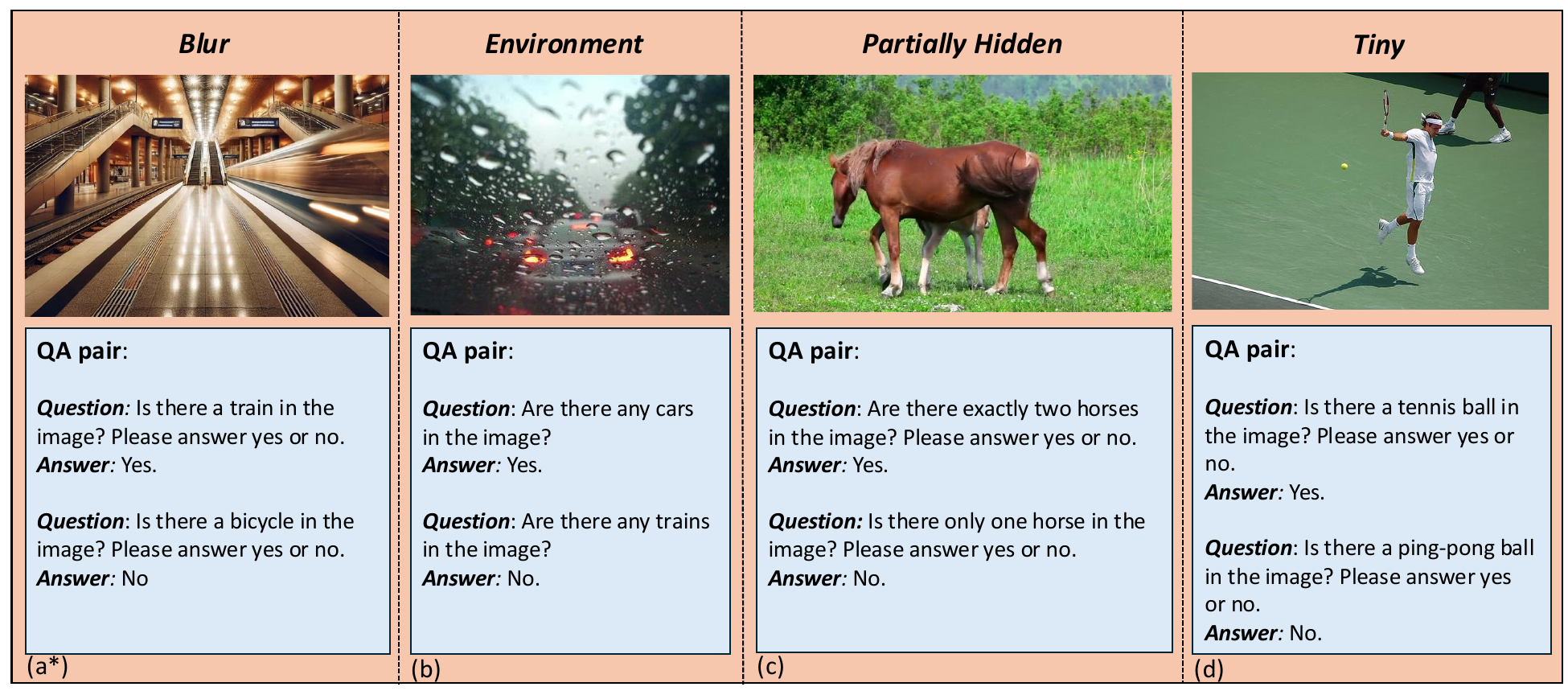}
\caption{Some examples of LanP. There are four categories in our benchmark: Blur, Environment, Partially Hidden, and Tiny. We show one sample figure and corresponding questions for each category. } 
\label{fig:example}
\end{figure*}

LanP is designed to evaluate the positive roles of language priors in LVLMs. Figure~\ref{fig:example} shows some examples from our benchmark. In this section, we first present the taxonomy of the dataset., followed by a description of the data collection process. Next, we will provide the details.

\begin{figure}[t!] 
\centering 
\includegraphics[width=0.8\columnwidth]{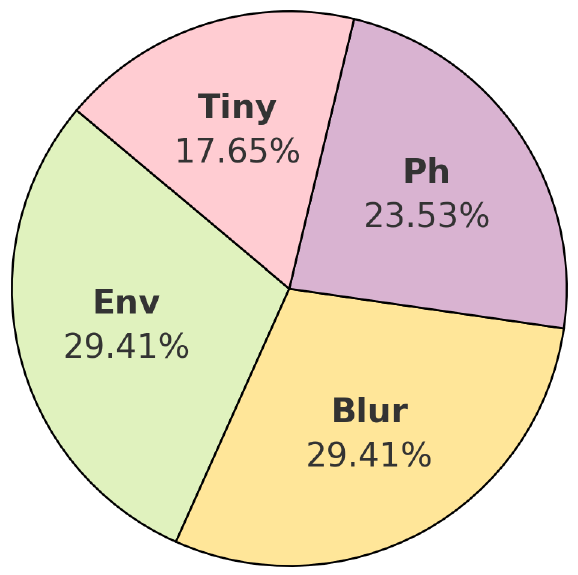}
\caption{Image Statistics of Four Categories in the LanP Dataset.} 
\label{fig:stat}
\end{figure}

\subsection{Overview of LanP}
\label{subsec:overview}
Our goal is to assess the positive role of language priors in LVLMs. Therefore, we selected questions where certain objects in the images are not very clear, challenging the models' ability to accurately interpret visual information, aiming to show that language priors can aid understanding when visual information is ambiguous. It could be hard for LVLMs to answer questions based only on visual information. Next, we introduce the details.

In our dataset, the image could provide enough background/environment information for the model. However, certain objects remain unclear due to factors such as darkness and blurriness, making it difficult for the model to interpret and output the answer correctly. If the model can better utilize language priors to understand the overall context of the image, it can provide more accurate answers. We consider four types of scenarios: \textit{blur}, \textit{environment}, \textit{partially hidden}, and \textit{tiny}. 

\textbf{\textit{Blur}} means some objects in the picture are blurred due to motion, out-of-focus, low resolution, and other reasons. LVLMs can better answer the questions by utilizing the background information.

Figure~\ref{fig:example} (a) is an illustration of this subcategory. The train is very blurry in the picture because it is moving. However, if the LVLMs can utilize other information in the picture, such as the platform and tracks, through language priors, it can help LVLMs provide a more accurate answer. Please note that this is not the original image that we included in our dataset but a similar image generated by ChatGPT due to licensing considerations\footnote{The original image of example (a) can be accessed at \url{https://www.cse.cuhk.edu.hk/~leojia/projects/dblurdetect/dataset.html} under the filename motion0031.jpg.}. 

\begin{table*}[h]

\small
    \centering
    \renewcommand{\arraystretch}{1.4} 
    \setlength{\tabcolsep}{10pt}
    \begin{tabular}{ l c c c}
        \toprule
        \rowcolor[gray]{0.9} \textbf{Model Name} & \textbf{Size} & \textbf{Base Language Model} & \textbf{Release Date} \\
        \midrule
        GPT-4 Turbo ~\citep{openai2023gpt} & - & - &2023 \\
        GPT-4o & - & - &2024 \\
        GPT-4o mini& - & - &2024 \\
        Gemini 1.5 Pro ~\citep{reid2024gemini}& - & - & 2024\\
        Gemini 1.5 Flash ~\citep{reid2024gemini} & - & - & 2024\\

        \midrule
 InternVL2.5-26B ~\cite{chen2024expanding}& \textasciitilde26B & internlm2\_5-20b-chat~\cite{chen2024far}&2024 \\
 InternVL2.5-8B~\cite{chen2024expanding} & \textasciitilde8B & internlm2\_5-7b-chat~\cite{chen2024far}&2024 \\
 InternVL2.5-4B ~\cite{chen2024expanding}& \textasciitilde4B &Qwen2.5-3B-Instruct~\cite{yang2024qwen2} &2024 \\
 InternVL2.5-2B~\cite{chen2024expanding}& \textasciitilde2B & internlm2\_5-1\_8b-chat~\cite{chen2024far} &2024 \\
 InternVL2.5-1B~\cite{chen2024expanding} & \textasciitilde1B & Qwen2.5-0.5B-Instruct~\cite{yang2024qwen2}&2024 \\
         \midrule
 InternVL2-26B~\cite{chen2024far}& \textasciitilde26B &internlm2-chat-20b~\cite{cai2024internlm2} &2024 \\
 InternVL2-8B~\cite{chen2024far} & \textasciitilde8B & internlm2\_5-7b-chat~\cite{cai2024internlm2}&2024 \\
 InternVL2-4B ~\cite{chen2024far}& \textasciitilde4B &Phi-3-mini~\cite{abdin2024phi} &2024 \\
 InternVL2-2B ~\cite{chen2024far}&\textasciitilde2B &internlm2-chat-1.8b~\cite{cai2024internlm2} &2024 \\
 InternVL2-1B ~\cite{chen2024far}& \textasciitilde1B & Qwen2-0.5B-Instruct~\cite{abs-2407-10671} &2024 \\
         \midrule
        Cambrian-13B~\cite{tong2024cambrian} & \textasciitilde
13B & Vicuna-1.5-13B~\cite{vicuna2023} &2024 \\
        Cambrian-8B~\cite{tong2024cambrian} & \textasciitilde
8B & Llama-3-Ins-8B~\cite{dubey2024llama} &2024\\
       \midrule
       Mini-InternVL-Chat-4B-V1-5 ~\cite{gao2024mini}& \textasciitilde
4B & Phi-3-mini~\cite{abdin2024phi}&2024 \\
        Mini-InternVL-Chat-2B-V1-5 ~\cite{gao2024mini}& \textasciitilde
2B &InternLM2-Chat-1.8B~\cite{cai2024internlm2} &2024\\
    
            \midrule
        LLaVA-NeXT-13B ~\cite{liu2024llavanext} & \textasciitilde
13B & Vicuna-13B~\cite{vicuna2023}&2024 \\
        LLaVA-NeXT-7B ~\cite{liu2024llavanext} & \textasciitilde
7B & Vicuna-7B~\cite{vicuna2023}&2024 \\
        \midrule
        LLaVA-1.5-13B ~\cite{liu2023improved} & \textasciitilde
13B & Vicuna-13B~\cite{vicuna2023}&2023 \\
        LLaVA-1.5-7B ~\cite{liu2023improved} & \textasciitilde
7B &Vicuna-7B~\cite{vicuna2023} &2023 \\

    \midrule
       ShareGPT4V-13B~\cite{chen2024sharegpt4v} & \textasciitilde
13B &Vicuna-13B~\cite{vicuna2023} & 2024\\
        ShareGPT4V-7B ~\cite{chen2024sharegpt4v} & \textasciitilde
7B & Vicuna-7B~\cite{vicuna2023}& 2024\\
        
        \bottomrule
    \end{tabular}
    \caption{Summary of Evaluated LVLMs.}
    \label{tab:lvlms}
    \label{tab:vlm_summary}
\end{table*}

\textbf{\textit{Environment}} refers to the difficulty of object recognition in the picture due to environmental factors such as rain, snow, dust storms, and darkness. Figure~\ref{fig:example} (b) is an illustration of this subcategory. There are some cars in the image, but they are not very clear due to the rain. If the model could better incorporate contextual information, such as the road and brake lights, it might be able to provide a more accurate response.

\textbf{\textit{Partially hidden}} refers to the situation in which the object we want to recognize in the picture is partially obscured by some objects, such as people sitting in cars and animals blocking each other; resulting in the inability to obtain all the characteristics of the target object and making it difficult to count the exact number of certain objects. Figure~\ref{fig:example} (c) is an illustration of this subcategory. We can see that there are two horses in the image, but one of them is mostly obscured by the horse closer to the camera. As a result, at first glance, it appears that there is only one horse. However, if the model can effectively utilize background information, such as the number of legs, it can arrive at the correct answer.

\textbf{\textit{Tiny}} means the objects in the image are very small, such as a tennis ball in a court. Figure~\ref{fig:example} (d) is an illustration of this subcategory. There is a small tennis ball in the image that has been hit by a racket. If the model can analyze background information in the image using language priors, such as the tennis racket and the tennis court, it will be able to answer the question more accurately.

\subsection{Dataset construction}
\label{subseq:dataset}
In this subsection, we introduce the details of dataset construction, including image collection and question design.

\subsubsection{Image collection}
We select the images that fit the requirements in Subsection~\ref{subsec:overview}. The parts of the image other than the queried object should ideally provide relevant background information, as shown in the example in Figure~\ref{fig:example}. 

We manually select 170 images from diverse sources and manually annotated them. The details of used data sources can be found in Appendix~\ref{app:ds}.

The \textit{Blur} and \textit{Environment} categories each contain 50 images, while the \textit{Partially Hidden} category has 40 images and the \textit{Tiny} category includes 30 images. The statistics of the dataset are also shown in Figure~\ref{fig:stat}.

\subsubsection{Question design}
\label{subsec:qd}

For questions, we focus on objects in images that are difficult to identify visually (e.g., blurred due to motion, obscured by other items) as the target of inquiry. To ensure the quality of the questions, each question is reviewed by another team member who is not the original designer of the question.

For Blur, Environment, and Tiny, the questions primarily address the issue of existence. Many questions in partially hidden explore the attributes such as the number of certain objects. For each image, we designed two paired questions where one is labeled as yes while another one is a negative sample with a label of no.

\section{Experiments}
In this section, we evaluate the positive impact of language priors in various LVLM.

\subsection{Experimental Settings}
\textbf{Selected LVLMs.}
To fully evaluate the quality of LanP and understand the effects of language priors in visual question answering, we conduct comprehensive experiments on 25 representative  closed-source and open-source LVLMs:
\begin{itemize}[leftmargin=*]
    \item \textit{Closed-source Models}: GPT-4 Turbo~\citep{openai2023gpt}, GPT-4o\footnote{\url{https://openai.com/index/hello-gpt-4o/}}, GPT-4o mini, Gemini 1.5 Pro~\citep{reid2024gemini}, and Gemini 1.5 Flash; 
    \item \textit{Open-source Models}: LLaVA-NeXT-7B/13B~\citep{liu2024llavanext}, LLaVA-1.5-7B/13B~\citep{liu2023improved}, ShareGPT4V-7B/13B~\citep{chen2024sharegpt4v}, Mini-InternVL-Chat-V1-5-2B/4B~\cite{gao2024mini}, InternVL2-1B/2B/4B/8B/26B, InternVL2.5-1B/2B/4B/8B/26B, Cambrian-8B/13B~\cite{tong2024cambrian}.
\end{itemize}
We also provide a summary of these models in Table~\ref{tab:vlm_summary}, including their model name, model size, base language model, and release date.

\begin{table*}[t]
\small
\centering
\renewcommand{\arraystretch}{1.2} 
\setlength{\tabcolsep}{6pt}
\caption{Performance of various LVLMs on the LanP Benchmark. The top five results in each category, as well as the overall performance, are highlighted in bold. ``Env'' stands for ``Environment'' and ``Ph'' stands for ``partially hidden''. Acc means accuracy.}
\label{tab:result_hard}
\resizebox{0.9\textwidth}{!}{
\begin{tabular}{l|c|c|c|c|c}
        \hline
        \textbf{Model} & Env Acc & Ph Acc & Blur Acc & Tiny Acc & Overall Acc \\
        \hline
\rowcolor{lightyellow} 
\multicolumn{6}{c}{\textbf{Closed-source (API-Based) Models} }\\ \hline
GPT-4 Turbo & 0.7800 & 0.4500& 0.7800 & 0.5333& 0.6588 \\
GPT-4o & 0.9000 & \textbf{0.7750} & \textbf{0.9400} & \textbf{0.8333} & \textbf{0.8706} \\
GPT-4o mini & 0.8200 & 0.6250 & \textbf{0.8800} & 0.5333 & 0.7412\\

Gemini 1.5 Pro & 0.9000 & \textbf{0.7250} & 0.8400 & \textbf{0.8333}  & \textbf{0.8294} \\
Gemini 1.5 Flash & \textbf{0.9400} & \textbf{0.7000} & \textbf{0.9200} & \textbf{0.8333} &  \textbf{0.8588}\\
\hline
\rowcolor{softblue} 
\multicolumn{6}{c}{\textbf{Open-source Models}} \\ \hline
InternVL2.5-26B & \textbf{0.9400} & \textbf{0.7500} & 0.8600 & \textbf{0.8667} & \textbf{0.8588} \\
InternVL2.5-8B & 0.9000 & 0.5500 & \textbf{0.9400} & \textbf{0.8333} & 0.8176 \\
InternVL2.5-4B & \textbf{0.9400} & 0.5500 & \textbf{0.9600} & \textbf{0.8667}  & \textbf{0.8412} \\
InternVL2.5-2B & \textbf{0.9400} & 0.4500 & \textbf{0.8800} &0.6667  & 0.7588 \\
InternVL2.5-1B & 0.8200 & 0.4000 & 0.8600 & 0.6667 & 0.7059 \\
\hline
InternVL2-26B & \textbf{0.9200} & 0.6250 & \textbf{0.8800} &0.7667  &0.8118\\
InternVL2-8B & 0.7800 & 0.4500 & 0.8000 &  0.7000& 0.6941 \\
InternVL2-4B & 0.8200 & 0.4500 & 0.7400 & 0.6667 & 0.6824 \\
InternVL2-2B & 0.7600 & 0.2500 & 0.7000 & 0.3333 & 0.5471 \\
InternVL2-1B & 0.7600 & 0.3500 & 0.8200 & 0.6000 & 0.6529 \\
\hline
Cambrian-13B & 0.8600 & 0.4750 & 0.7800 & 0.6333 & 0.7059 \\
Cambrian-8B &  0.9000 & \textbf{0.7000} & 0.8400 & 0.7667 & 0.8118\\
\hline
Mini-InternVL-Chat-4B-V1-5& 0.7800 & 0.4000 & 0.8000 & 0.7000 &  0.6824\\
Mini-InternVL-Chat-2B-V1-5& 0.8800 & 0.2000 & 0.8200 & 0.6000 & 0.6529 \\
\hline
LLaVA-NeXT-13B & 0.8600 & 0.5250 & 0.7800 & 0.7000 & 0.7294 \\
LLaVA-NeXT-7B & 0.8800 & 0.4500 & 0.7400 &  0.7000& 0.7059 \\
\hline
LLaVA-v1.5-13B & 0.8200 & 0.4250 & 0.8000 & 0.7333 & 0.7059 \\
LLaVA-v1.5-7B & 0.6800 & 0.3750 & 0.5800 & 0.6000 & 0.5647 \\
\hline
ShareGPT4V-13B & 0.8400 & 0.2750 & 0.8600 &0.6667 & 0.6824 \\
ShareGPT4V-7B & 0.8200 &0.3250 & 0.5800 & 0.6333 & 0.6000 \\
\hline
\end{tabular}
}
\end{table*}

\noindent\textbf{Evaluation Metrics.} We use \textit{\textbf{Accuracy}} as our evaluation metrics. Accuracy is the most widely used evaluation indicator~\citep{LiDZWZW23,huang2024visual,liu2024phd}. In our benchmark, each image has two questions where one is labeled as yes and another one is labeled as no as shown in Figure~\ref{fig:example}. We require that \textit{only both questions are answered correctly at the same time can be counted as correct}.

Following the settings in VLMEvalKit~\citep{duan2024vlmevalkit}, we extract the answer of tested LVLMs based on their output containing Yes or No. To be specific, if the model's response contains "Yes" but not "No," the output will be extracted as "Yes." Conversely, if it contains "No" but not "Yes," the output will be extracted as "No."

To prevent cases where the model's output contains both "Yes" and "No," GPT will be used to determine the final output in such situations.

\noindent\textbf{Configuration.} Our code is based on VLMEvalKit~\citep{duan2024vlmevalkit}. All experiments are conducted on machines with four Nvidia RTX A6000 GPUs, each with 48 GB of memory.

\subsection{Experimental Results}

In this subsection, we test the performance of various LVLMs.

\subsubsection{Overall Results} 
The overall results of our benchmark are shown in Table~\ref{tab:result_hard}. 

We can observe that all the models have an overall accuracy greater than 0.5. A major reason for their high overall accuracy is their good performance in the Environment, Blur, and Tiny categories. The questions in these categories primarily focus on existence problems, which are relatively easier compared to those in the partially hidden category.

We can see that GPT-4o achieves the highest overall accuracy of 0.8706. Most closed-source models achieve satisfactory performance except GPT-4 Turbo. To our surprise, GPT-4 Turbo has a much lower accuracy compared with other closed-source models. For example, GPT-4o mini achieves an overall accuracy of 0.7412, which is higher than the 0.6588 accuracy obtained by GPT-4 Turbo. 

In general, closed-source models show better overall performance compared with open-source models. However, some large open-source models also achieve very good performance. For instance, we can observe that InternVL2.5-26B achieves the highest overall performance among open-source models, with an accuracy of 0.8588. This accuracy is also the second highest among all LVLMs, matching that of Gemini 1.5 Flash.

\subsubsection{Performance on Env, Blur, and Tiny Parts} 

We observe that many models achieve very high accuracy in these three parts, especially in the Environment and Blur parts. For example, GPT-4o achieves the highest accuracy of 0.9400 in the Blur category, while the lowest accuracy 0.5800 obtained by ShareGPT4V-7B and LLaVA-v1.5-7B is also higher than 0.5.

Therefore, we can conclude that the existing newer LVLMs, whether open-source or closed-source models, can effectively utilize relevant information in images to address object existence questions.

\subsubsection{Performance on Partially Hidden Part}
Accuracy on our \textit{partially hidden} task is generally much lower than in the other three categories, underscoring the challenge this task presents for current LVLMs. Although a few models achieve satisfactory performance, the majority of LVLMs exhibit an accuracy below 0.5; for instance, even GPT-4 Turbo achieves only 0.45. 

\begin{figure*}
\centering 
\includegraphics[width=1.65\columnwidth]{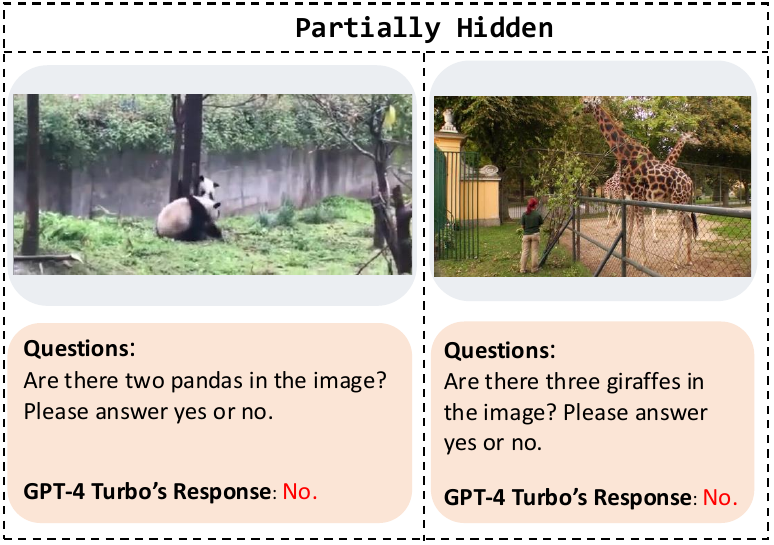}
\caption{Some questions that GPT-4 Turbo answers incorrectly.} 
\label{fig:failure}

\end{figure*}

We also show some failure cases of GPT-4 Turbo in Figure~\ref{fig:failure}. We can see that these questions, which are very simple for humans, cannot be correctly answered by GPT-4 Turbo. If the model could better utilize contextual information, such as the number of giraffe legs in the right image, it should be able to answer such questions more accurately.

Our results suggest that although many models can leverage contextual visual information to address the existence problem, they struggle to accurately count the objects when they are partially obscured by others. For example, Mini-InternVL-Chat-2B-V1-5 demonstrates a high accuracy of 0.88 in the environment part. However, its accuracy drops to just 0.2 in the partially hidden part. 

\subsubsection{Impact of Model Size}
We can first look at the results within the same model family (Here we refer to models share the same LLM series). Generally, models incorporating a larger language model tend to achieve better overall performance. For example, InternVL2.5-8B achieves higher overall accuracy compared to its smaller counterpart InternVL2.5-2B. Similar trends are also observed in the LLaVA-1.5 series models where LLaVA-1.5-13B performs better than LLaVA-1.5-7B in all tasks. 

Note that InternVL2.5-4B and InternVL2.5-1B utilize a different language model architecture than InternVL2.5-8B, as shown in Table~\ref{tab:lvlms}.

Assuming that the larger the language model is, the more world knowledge it possesses and the stronger its language prior. Then, when language prior is needed to analyze scene information in an image, a stronger language prior can perform better analysis, leading to better results. Language priors play a crucial role in helping LVLMs better interpret visual information.

\section{Conclusion}
In this work, we introduce LanP, a novel benchmark designed to reassess the role of language priors in large vision-language models. LanP comprises 170 images and 340 carefully crafted questions, requiring models to effectively utilize language priors to give answers.
 
Extensive experiments on 25 representative LVLMs demonstrate that current models still struggle with the Partially hidden part in LanP, achieving unsatisfactory performance. This suggests that the language priors of many existing models are not strong enough to handle such challenging cases in LanP.

For many LVLM series, increasing the size of the underlying LLM generally leads to higher accuracy because large LLMs have stronger language priors and have more world knowledge. Our study suggests that we should find a good balance where language priors enhance responses by providing beneficial world knowledge while avoiding the risk of overriding visual information and causing hallucinations. We anticipate that LanP could play a valuable role in the development and evaluation of stronger large vision-language models with more robust language priors in the future.

\section{Limitations}
Considering the issue of runtime and ease of evaluation, following POPE~\citep{LiDZWZW23}, our benchmark prompts LVLMs to answer yes or no to evaluate the model. However, this approach might not fully reflect the model's ability to generate more free-form responses in various scenarios. One future direction is to remove the yes/no constraint and explore the performance difference.

Another limitation is that there is some overlap between categories. For instance, a blurry image could potentially also be classified as an environmental image (e.g., a dark scene). In such cases, we will randomly assign the image to one of the applicable categories.

\bibliography{custom}

\appendix

\newpage

\section{Data source}
\label{app:ds}
\begin{itemize}
  \item \textbf{Photos taken by the authors}

    \item \textbf{MPII Human Pose Dataset}~\citep{andriluka14cvpr,pishchulin2014fine}: The MPII Human Pose Dataset was downloaded from \url{http://human-pose.mpi-inf.mpg.de/}. The dataset provides annotated human poses in various activities.

    \item \textbf{Blur Detection Dataset}~\citep{shi2014discriminative}: The Blur Detection Dataset can be found at \url{https://www.cse.cuhk.edu.hk/~leojia/projects/dblurdetect/dataset.html}. It includes images with different levels of blur, making it useful for developing and testing blur detection algorithms.

    \item \textbf{Stereo Blur Dataset}~\citep{Zhou2019Stereodeblur}: It contains stereo image pairs with different blur levels, designed for stereo vision and blur detection. The data can be obtained from this site: \url{https://shangchenzhou.com/projects/stereoblur/}.

    \item \textbf{Weather Phenomenon Database (WEAPD)}~\citep{M8JQCR2021}: The Weather Phenomenon Database was sourced from \url{https://dataverse.harvard.edu/dataset.xhtml?persistentId=doi:10.7910/DVN/M8JQCR}. This database contains comprehensive records of various weather phenomena, used for atmospheric and environmental studies.

    \item \textbf{OVIS}~\citep{qi2022occluded}: The OVIS dataset was downloaded from \url{https://songbai.site/ovis/}. This dataset provides videos of objects under occlusions for testing object-tracking algorithms.

    \item \textbf{ExDark}~\citep{Exdark}: The ExDark dataset was downloaded from \url{https://github.com/cs-chan/Exclusively-Dark-Image-Dataset}. This dataset contains images captured under different lighting conditions, useful for low-light image processing research.

    \item \textbf{The Visual Genome Dataset}~\citep{krishna2017visual}: The Visual Genome Dataset was accessed from \url{https://homes.cs.washington.edu/~ranjay/visualgenome/api.html}. This dataset provides a detailed visual knowledge base with descriptions and annotations.

\item \textbf{MS COCO}~\citep{LinMBHPRDZ14}
: MS COCO is a widely used datasets that contains more than 320K images.

\end{itemize}

\end{document}